\documentclass[lettersize,journal]{IEEEtran}
\usepackage{amsmath,amsfonts}
\usepackage{algorithmic}
\usepackage{algorithm}
\usepackage{array}
\usepackage{textcomp}
\usepackage{stfloats}
\usepackage{url}
\usepackage{verbatim}
\usepackage{graphicx}
\usepackage{cite}
\usepackage{color,soul}
\usepackage{lipsum}
\usepackage{graphicx}
\usepackage{caption}
\usepackage{subcaption}
\usepackage{amsmath}
\usepackage{hyperref}

\usepackage[table,xcdraw]{xcolor}

\definecolor{Gray}{gray}{0.88}
\definecolor{DarkGray}{gray}{0.8}
\definecolor{LightGray}{gray}{0.93}

\begin{document}


\title{A POV-based Highway Vehicle Trajectory Dataset and Prediction Architecture}

\author{
    \IEEEauthorblockN{
    Vinit Katariya\IEEEauthorrefmark{1}, ~\IEEEmembership{Student Member},
    Ghazal Alinezhad~Noghre\IEEEauthorrefmark{1}, ~\IEEEmembership{Student Member},
    Armin Danesh~Pazho\IEEEauthorrefmark{1}, ~\IEEEmembership{Student Member},
    Hamed Tabkhi\IEEEauthorrefmark{1}, \IEEEmembership{Member,~IEEE}
    }

\thanks{
    \IEEEauthorrefmark{1} {Electrical and Computer Engineering Department, University of North Carolina Charlotte, Charlotte,
	NC, 28223 USA  (e-mail:  vkatariy@uncc.edu, galinezh@uncc.edu, adaneshp@uncc.edu ,  htabkhiv@uncc.edu).}
	}
}



\maketitle

\begin{abstract}

Vehicle Trajectory datasets that provide multiple point-of-views (POVs) can be valuable for various traffic safety and management applications. Despite the abundance of trajectory datasets, few offer a comprehensive and diverse range of driving scenes, capturing multiple viewpoints of various highway layouts, merging lanes, and configurations. This limits their ability to capture the nuanced interactions between drivers, vehicles, and the roadway infrastructure. We introduce the \emph{Carolinas Highway Dataset (CHD\footnote{\emph{CHD} available at: \url{https://github.com/TeCSAR-UNCC/Carolinas\_Dataset}})}, a vehicle trajectory, detection, and tracking dataset. \emph{CHD} is a collection of 1.6 million frames captured in highway-based videos from eye-level and high-angle POVs at eight locations across Carolinas with 338,000 vehicle trajectories. The locations, timing of recordings, and camera angles were carefully selected to capture various road geometries, traffic patterns, lighting conditions, and driving behaviors.

We also present \emph{PishguVe}\footnote{\emph{PishguVe} code available at: \url{https://github.com/TeCSAR-UNCC/PishguVe}}, a novel vehicle trajectory prediction architecture that uses attention-based graph isomorphism and convolutional neural networks. The results demonstrate that \emph{PishguVe} outperforms existing algorithms to become the new state-of-the-art (SotA) in bird's-eye, eye-level, and high-angle POV trajectory datasets. Specifically, it achieves a 12.50\% and 10.20\% improvement in ADE and FDE, respectively, over the current SotA on NGSIM dataset. Compared to best-performing models on CHD, \emph{PishguVe} achieves lower ADE and FDE on eye-level data by  14.58\% and 27.38\%, respectively, and improves ADE and FDE on high-angle data by 8.3\% and 6.9\%, respectively.
\end{abstract}


\section{Introduction}
Intelligent Transportation Systems (ITS) have become integral to modern transportation networks, leveraging various technologies to improve road safety and efficiency. One key aspect of ITS is using trajectory prediction networks, which can accurately forecast the movements of vehicles, pedestrians, and other road users \cite{wang2020multi, mo2022multi, cai2021pedestrian, mozaffari2020deep}. In recent years, advancements in technology, such as accident detection systems \cite{tian2019accidentDetect,singh2019accidentDetect}, traffic surveillance systems \cite{zhang2021survailanceSystem}, and lane departure warning systems \cite{Olofsson2021laneDeparture} have shown great potential in reducing accidents and saving lives on the road. However, highway safety has historically received less attention and innovation than other areas, possibly due to high costs and regulatory hurdles.

The National Highway Traffic Safety Administration reported 42,000 fatalities in 2022 \cite{NHTSA2022} due to highway-related motor accidents. In addition, 857 fatalities were reported in 2020 \cite{FHWA2023} in work zone accidents, with a 4\% increase in the death toll over the past two years \cite{AGCA2022}. These statistics demonstrate a significant scope for improvement in ITS applications related to highway surveillance and safety.

Vehicle trajectory datasets are essential for studying traffic behavior and analyzing macroscopic traffic data\cite{tian2019accidentDetect, Sochor2019VehSpeed}. However, most available datasets provide only bird's-eye or dashcam views \cite{NGSIM_i80, NGSIM_US101, highDdataset, agroverse, lyft, waymo} or low-quality high-angle view videos\cite{shah2018cadp, ren2018trafficBehav}. These perspectives may be insufficient for highway safety and surveillance applications\cite{ruimini2017Bidirectional}, mainly when the environment is subject to variation. Eye-level and high-angle views of highway traffic with merging lanes are particularly important for highway-based safety applications, such as work zone safety\cite{sabeti2021toward}, collision avoidance\cite{Cheng2021collisionAvoid } and lane departure warning systems\cite{Olofsson2021laneDeparture}, as well as surveillance applications, such as traffic monitoring and incident detection\cite{tian2019accidentDetect}. However, they are not typically found in trajectory datasets. High-angle view data, for example, can be used for incident detection and response, traffic monitoring\cite{HuangHRR20II,ren2018trafficBehav} and analysis\cite{ren2018trafficBehav,Clausse2019extraction, wang2019towards}, which can be challenging to obtain in real-time with bird's-eye view data. Similarly, Eye-level view data can be helpful for real-time pedestrian and worker safety and alarm applications\cite{ren2020cooperative, katariya2022deeptrack1, sabeti2021toward} for work zones and highway management. 

While many trajectory datasets have served as crucial benchmarks in their respective focus areas, they often lack multiple POVs of incoming traffic. They rarely examine the vehicle dynamics during lane merges which is essential in comprehending the naturalistic driving maneuvers close to the real-world worksite. This leaves a blind spot for research on highway-based edge ITS applications involving transportation safety and AI. 

In light of this, we propose the \emph{Carolinas Vehicle Dataset (CHD)}, a comprehensive trajectory dataset capturing naturalistic highway driving behavior from multi-POVs. CHD consists of 338,000 unique trajectories with five highway vehicle categories, varying geometries, and lane mergers. It also provides 1.6 million high-resolution frames with high traffic density recorded during varying lighting and weather conditions. Thus, CHD includes annotations and trajectory information for highway vehicles, making it a valuable resource for ITS and computer vision applications to advance the development of safer and more efficient transportation systems.

\begin{table*}[t]
\centering
\caption{Comparison of existing trajectory datasets with CHD. T and BB in the Annotation column stand for Trajectories and Bounding Boxes respectively. FPS stands for Frames Per Second.}
\label{tab:traj_summ}
\resizebox{\textwidth}{!}{%
\begin{tabular}{c||c|c|c|c|c|c|c|c}
\rowcolor{DarkGray}
Dataset            & \begin{tabular}[c]{@{}c@{}}Length \\ (h)\end{tabular} & Frames & \begin{tabular}[c]{@{}c@{}}Bounding\\ Boxes\end{tabular} & \begin{tabular}[c]{@{}c@{}}Sampling\\ Rate (FPS)\end{tabular} & \begin{tabular}[c]{@{}c@{}}Lanes\\ Detected\end{tabular} & \begin{tabular}[c]{@{}c@{}}Localities\\ Covered\end{tabular} & Annot. & \begin{tabular}[c]{@{}c@{}}POV\end{tabular}                       \\ \hline \hline
NGSIM\cite{NGSIM_i80, NGSIM_US101}             & 1.5        & 11.2K      & -                                                        & 10                                                      & Yes                                                      & 2                                                    & T           & Bird's-eye                                                                   \\
HighD\cite{Krajewski2018highD}            & 16.5       & -      & -                                                        & 25                                                      & Yes                                                      & 6                                                    & T           & Bird's-eye                                                                   \\
Agroverse2 Fore. \cite{Wilson2021Agro}  & 763        & -   & -                                                     & 10                                                        & No                                                       & 2                                                   & T           & Dashcam                                                                    \\
AppolloScape\cite{ma2019trafficpredict} & 100       & 93K     & 81.8K                                                        & 10                                                        & No                                                       & 4                                                   & T           & Dashcam                                                                    \\
Lyft\cite{Houston2020Lyft}               & 118       & 232K      & 1.3M                                                        & 10                                                        & No                                                       & 1                                                   & T           & Dashcam                                                                    \\
WAYMO\cite{ettinger2021WAYMO}            & 5.5      & 1M      & 12M                                                        & 10                                                        & No                                                        & 3                                                   & 3D BB, T           & Dashcam                                                                           \\
nuScenes\cite{Caesar2020nuScenes}           & 5.5       & 1.4M   & 1.4M                                                     & 2                                                      & No                                                       & 2                                                   & 3D BB, T    & Dashcam                                                                    \\ \rowcolor{Gray} \hline
\textbf{CHD (Ours)}               & 7.5        & 1.6M   & 1.39M$^*$                                                      & 5, 60                                                     & No                                                       & 4                                                   & 2D BB, T    & \begin{tabular}[c]{@{}c@{}}High-angle\\ Eye-level\end{tabular}  \\

\multicolumn{9}{r}{$^*$\scriptsize{\textit{Bounding boxes in CHD's trajectory data.}}}

\end{tabular}%
}
\vspace{-15pt}
\end{table*}

We also introduce \emph{PishguVe}, a trajectory prediction architecture that leverages attention for graph isomorphism and convolution neural networks to achieve SotA performance across datasets with multiple POVs. PishguVe provides a solution to overcome the limitations of POV-based trajectory prediction approaches, enabling more generalizable models for real-world applications. Compared to existing approaches evaluated on widely used NGSIM dataset, PisguVe surpasses current SotA \cite{noghre2022pishgu} with 12.50\% and 10.20\% improvement over ADE and FDE, respectively. It also outperforms existing approaches in ADE and FDE by 14.58\% and 27.38\% when evaluated on CHD eye-level and by 8.3\% and 6.9\%, respectively, on CHD high-angle POV dataset.

In summary, the contributions of this article are:
\begin{itemize}
\item We introduce the \textit{CHD}, a vehicle trajectory dataset from multiple POVs capturing vehicle maneuvers on highways with varying geometries and lane mergers. CHD also provides 1.6 million high-resolution images with annotation data for detection and tracking applications.

\item We introduce \textit{PishguVe}, a SotA trajectory prediction network based on attention-based convolutional and graph isomorphism networks. PishguVe achieves a best-in-class error rate when evaluated on multiple datasets, making it well-suited for real-time trajectory prediction in different POV scenarios.

\item To verify the effectiveness of \textit{CHD} and \textit{PishguVe}, we present a comprehensive comparison of the trajectory dataset and assessment of trajectory prediction algorithms trained and evaluated on NGSIM and CHD in this manuscript.
\end{itemize}

\section{Related Works}
\subsection{Vehicle Tracking and Trajectory Datasets}

NGSIM dataset \cite{NGSIM_i80, NGSIM_US101} is one of the most extensive available datasets recorded in a bird's-eye view angle. This dataset is collected from two freeways, namely I-80 and US-101. This dataset provides trajectories extracted at a rate of 10Hz. HighD dataset \cite{highDdataset} is recorded in six locations on German highways using drones at the bird's-eye POV. However, the original videos are not provided; only the extracted trajectories using computer vision algorithms are available. Agroverse 2 Motion Forecasting dataset \cite{agroverse} provides trajectory data for different classes of vehicles and pedestrians extracted from videos recorded in six cities. All the mentioned datasets are collected from a bird's-eye POV. However, most of the time, bird's-eye view videos are unavailable in real-world scenarios since traffic cameras are usually mounted on top of the traffic lights. Another group of datasets, such as ApolloScape Trajectory \cite{apollo}, Lyft \cite{lyft}, WAYMO \cite{waymo}, and nuScenes \cite{nuscene} are specifically designed for autonomous vehicle applications and are recorded from a dashcam POV. These datasets are useful as benchmarks for trajectory prediction for several applications. However, their applicability to real-world highway safety applications is limited due to the properties of the data, such as moving cameras and viewpoints. Table \ref{tab:traj_summ} shows a detailed dataset analysis. 

For vehicle detection/tracking, some existing datasets with high-angle POV are listed in \ref{tab:cam_view}. The TRANCOS \cite{trancos} dataset is recorded at the high-angle POV and is a vehicle counting dataset captured using publicly available surveillance cameras in Spain. Unlike bird's-eye POV datasets, in TRANCOS, vehicles are highly overlapping and close to real-world scenarios. TRAF object tracking dataset \cite{traf} provides videos in high-angle and dashcam POVs. The UA-DETRA dataset \cite{detra1, detra2, detra3} is a challenging real-world dataset captured from a high-angle point of view. It consists of videos captured from 24 locations in China and is manually annotated. Although the dataset is introduced for multi-object tracking and detection, it can also be used as a benchmark for trajectory prediction.

Furthermore, a set of datasets mainly focused on accident and traffic behaviors. TCP \cite{ren2018learning} introduces a high-angle POV dataset, capturing a four-way intersection at different times of the day. It uses an object detector for labeling the vehicles, and it has hand labels for when the vehicle enters the intersection. CADP \cite{shah2018cadp} is another interesting dataset on car accidents. This dataset is made of videos from YouTube\footnote{www.youtube.com}. Because the type of collection inherently consists of various qualities, weather conditions, time of day, etc. A two-step labeling including hand-extracted start and end time and performing a spatiotemporal annotation using VATIC \cite{vondrick2013efficiently} creates the annotations. CAD-CVIS \cite{tian2019accidentDetect} is another dataset collected from video-sharing websites. It uses LabelImg \cite{heartexlabs} to localize the accident in the frames of the videos.

\begin{table*}[t]
\centering
\caption{Comparison of existing camera view vehicle detection and tracking datasets with CHD. In the function column, D is for Detection, TP is for Trajectory Prediction, AD is for Accident Detection, and T is for Tracking. FPS stands for Frames Per Second}
\label{tab:cam_view}
\resizebox{\textwidth}{!}{%
\begin{tabular}{c||c|c|c|c|c|c|c|c|c|c}
\rowcolor{DarkGray}
\textbf{Year} & \textbf{Dataset} & \textbf{Scenes} & \textbf{Length} & \textbf{Frames} & \textbf{2D Boxes} & \textbf{Class} & \textbf{POV} & \textbf{Resolution} & \textbf{FPS} & \textbf{Function} \\ \hline \hline
TRANCOS\cite{Olemedo2015TRANCOS}        & 2016          & 9               & -               & 1244            & 46.79K            & -              & High-angle            & 360p                & -            & D                     \\
TRAF \cite{Chandra2019TRAF}            & 2019          & 20              & 39m             & 12.4K               & 19.88K                 & 7              & High-Angle, Dashcam                     & 720p                & 10           & TP                      \\ 
CADP\cite{shah2018cadp}       & 2019          & -              & 5.2h             & 518K            & -             & 6              & High-angle            & Various               & -           & AD                   \\
CAD-CVIS\cite{tian2019accidentDetect}       & 2020          & 24              & -             & 228K            & -             & -              & High-angle            & Various                & -           & AD \\
UA-DETRA\cite{Wen2020UADETRAC}       & 2020          & 24              & 10h             & 140K            & 1.2M              & 2              & High-angle            & 540p                & 25           & D,T                 \\\rowcolor{Gray} \hline
\textbf{CHD (Ours)}             & 2023          & 8               & 7.5h           & 1.6M            &   33.47M$^*$                & 5              & Eye-level, High-angle & 1080p               & 60           & TP, D, T   \\

\multicolumn{11}{r}{$^*$\scriptsize{\textit{Bounding box annotations across all the recorded data in CHD.}}}

\end{tabular}%
}
\vspace{-15pt}
\end{table*}

\subsection{Vehicle Trajectory Prediction Algorithms}
Earlier models for trajectory prediction mostly used Recurrent Neural Networks such as Gated Recurrent Unit (GRU) or Long-short-term Memory (LSTM) for modeling the time dimension. CS-LSTM \cite{deo2018convolutional} has an LSTM-based encoder-decoder structure for embedding the vehicles' previous motions. On top of that, for modeling the interdependencies between vehicles, it uses a convolutional social pooling mechanism. GRIP++ \cite{li2019grip} uses graphs to model the interactions between vehicles, and, using an LSTM encoder-decoder, it predicts the future trajectory. Later, attention mechanisms found their way to trajectory prediction; using the Spatial-temporal Attention mechanism and LSTM units, STA-LSTM \cite{lin2021vehicle} improved the performance and explainability of trajectory prediction models. Trajectron++ \cite{salzmann2020trajectron} also has a recurrent structure based on LSTM and uses spatiotemporal graphs to model the input trajectories for vehicles and pedestrians. Like Trajectron++, Social-STGCNN \cite{mohamed2020social} takes advantage of spatiotemporal graphs, but instead of the recurrent neural networks, it uses a convolutional structure and is primarily designed for pedestrian trajectory prediction. 

DeepTrack\cite{katariya2022deeptrack1} focused on real-world trajectory prediction applications such as traffic management and introduced a lightweight and agile model capable of real-time inference. Unlike previous methods, DeepTrack uses Temporal Convolutional Networks (TCNs) for modeling the time dimension. With Cyber-physical Systems (CPS) applications in mind, Pishgu \cite{noghre2022pishgu} presents an efficient universal network architecture for trajectory prediction in different domains using Graph Isomorphism Networks (GINs) and convolutional attention mechanism. It is conventional for probabilistic trajectory prediction models to predict N number of possible future trajectories and pick the best one \cite{deo2018convolutional, mohamed2020social}. However, multiple predicted trajectories per subject in real-world scenarios are not very effective. Thus, more recent models such as \cite{li2019grip, lin2021vehicle, salzmann2020trajectron, katariya2022deeptrack1, noghre2022pishgu} primarily focus on single future trajectory prediction.

\begin{figure*}[htbp]
  \centering
  \includegraphics[width=.9\textwidth]{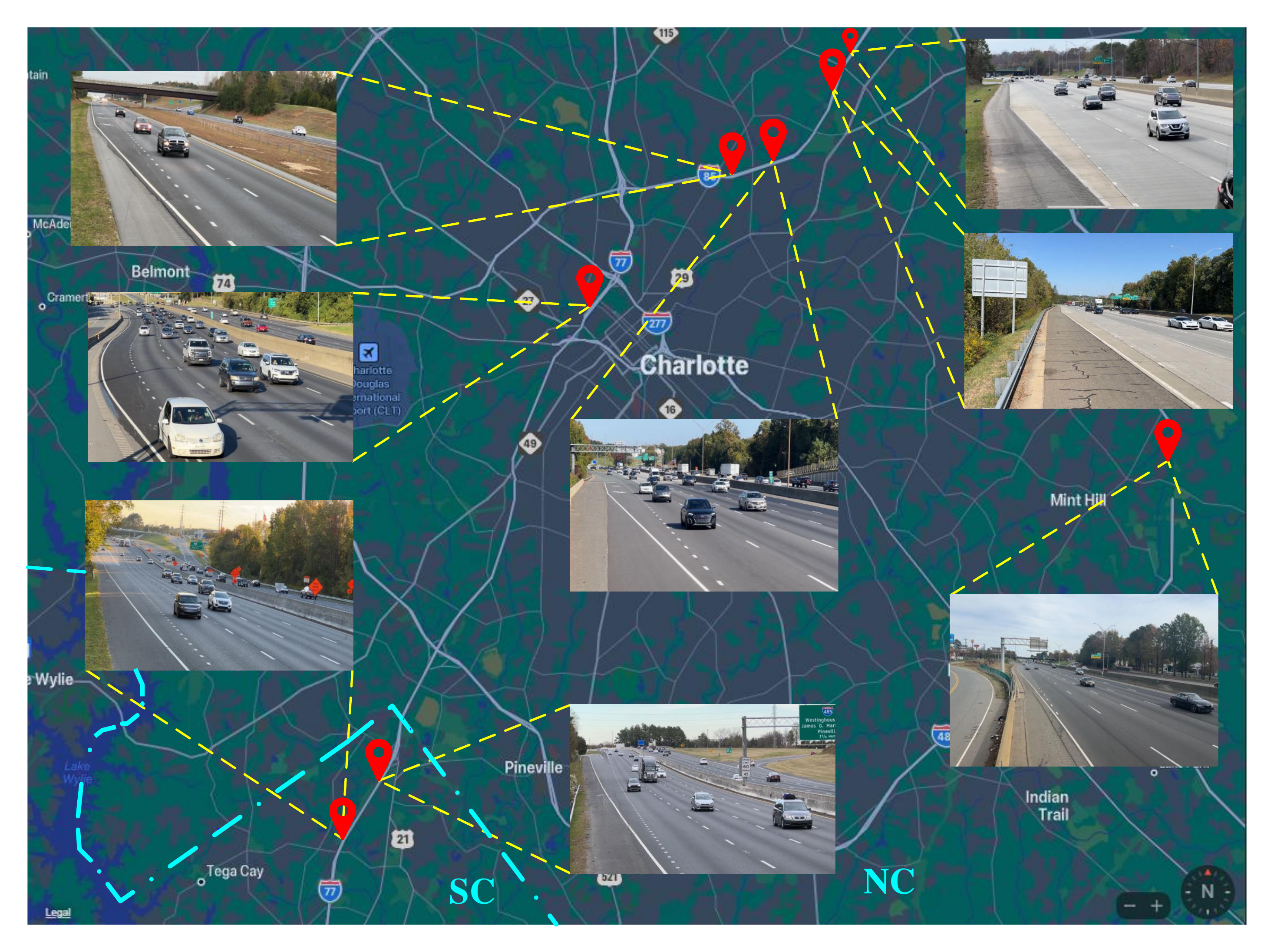}
  \vspace{-5pt}
  \caption{Spatial distribution of all the eight recording locations on the highways of NC and SC included in CHD.}
  \vspace{-5pt}
  \label{fig:POV_figures_all}
\end{figure*}
\vspace{-15pt}

\section{Carolinas Highway Dataset (CHD)}

CHD is a comprehensive collection of vehicle highway trajectories extracted from videos recorded across North Carolina and South Carolina from two distinct POVs: eye-level and high-angle. CHD also provides annotations for multi-vehicle detection and tracking. Fig. \ref{fig:POV_figures_all} shows the spatial map of all the recording locations with high-angle POV images of each site. The videos were recorded in full HD (1080p) resolution between 9 AM and 7 PM at eight locations within different localities.

The recording locations were selected to capture different traffic flows, vehicle behaviors, naturalistic driving patterns on highways with different road geometries, and varying lengths and structures of merging lanes. The videos were recorded at different times of the day to capture various lighting conditions, traffic patterns, and volume. CHD captures data from multiple POVs, providing a diverse perspective of incoming traffic. 


\subsection{Annotations and Trajectory Extraction}
\label{collection_extraction}

One of the unique features of the proposed dataset, CHD, lies in the use of generated annotations, which reflect the annotations available in real-world applications where hand annotations are often unavailable \cite{pazho2022chad}. Generated annotations may contain some noise or inaccuracies, but they provide a more realistic representation of annotations used in the real world. Using generated annotations, CHD can more accurately reflect the challenges in real-world applications, such as occlusions, varying lighting conditions, and other factors affecting detection and tracking accuracy.

CHD employs bounding box annotations to identify vehicles in each frame and spatially locate them. To generate high-quality bounding boxes, the CHD utilizes YOLOv5 \cite{glenn_jocher_2020_yolov5}. Specifically, the YOLOv5x6 model has been trained on the BDD100k dataset \cite{Yi2020bdd100k} to identify different types of vehicles.
 
 These bounding boxes provide crucial information that the tracking model uses to assign a unique identification number to each vehicle across multiple frames. CHD uses ByteTrack \cite{zhang2022bytetrack}, in combination with YoloV5 tracks multiple vehicles in a frame. ByteTrack, with a bounding box as a basic tracking unit, uses data association to match the vehicles in different frames assigning them a unique ID.  

In addition to utilizing the bounding box annotations to identify and locate vehicles in a given scene, the center of the bounding box is used to determine the position of each vehicle in the scene. This information is used to generate trajectories, which refer to the path followed by the object over a period of time. Specifically, in the context of vehicle tracking in CHD, the trajectories were generated by tracking the vehicle's position in successive video frames, using a unique ID assigned to each vehicle by ByteTrack as a unique trajectory in the dataset.

Several filters were applied to extracted raw trajectory data to ensure the accuracy of extracted trajectories. Unique trajectories of a minimum duration of four seconds and above were included in the dataset, enabling CHD for real-world models with smaller input and output windows. Additionally, stationary vehicles and vehicles moving away from the camera were filtered out, as the focus was on incoming traffic. False detections were also manually filtered out whenever possible to eliminate additional noise in the data.
 

 

\subsection{CHD Statistics and Format}
\label{subsec:stats}
CHD consists of 338,000 trajectories extracted from 16 videos, with a total of 1.6 million frames. This is outlined in Table \ref{tab:traj_summ}, demonstrating the scale of CHD comparable to that of well-known trajectory datasets. Regarding the number of bounding boxes for trajectory data, CHD aligns with widely used datasets, as shown in Table \ref{tab:traj_summ}. Along with trajectory data, it includes raw videos and bounding box annotations for all the recorded videos. The 33M bounding box annotations reported in Table \ref{tab:cam_view} showcase its high traffic density of recording environments.

In addition to the vast number of frames, CHD benefits from high-quality (1080p resolution) image data recorded at the frame rate of 60 fps, which exceeds others, as demonstrated in Table \ref{tab:cam_view}. As summarized in previous sections and Tables \ref{tab:traj_summ} and \ref{tab:cam_view}, CHD is among the few datasets with multiple POVs of incoming vehicles trajectory. 

\begin{figure}[]
\centering
\includegraphics[width=8cm, height= 5.5cm]{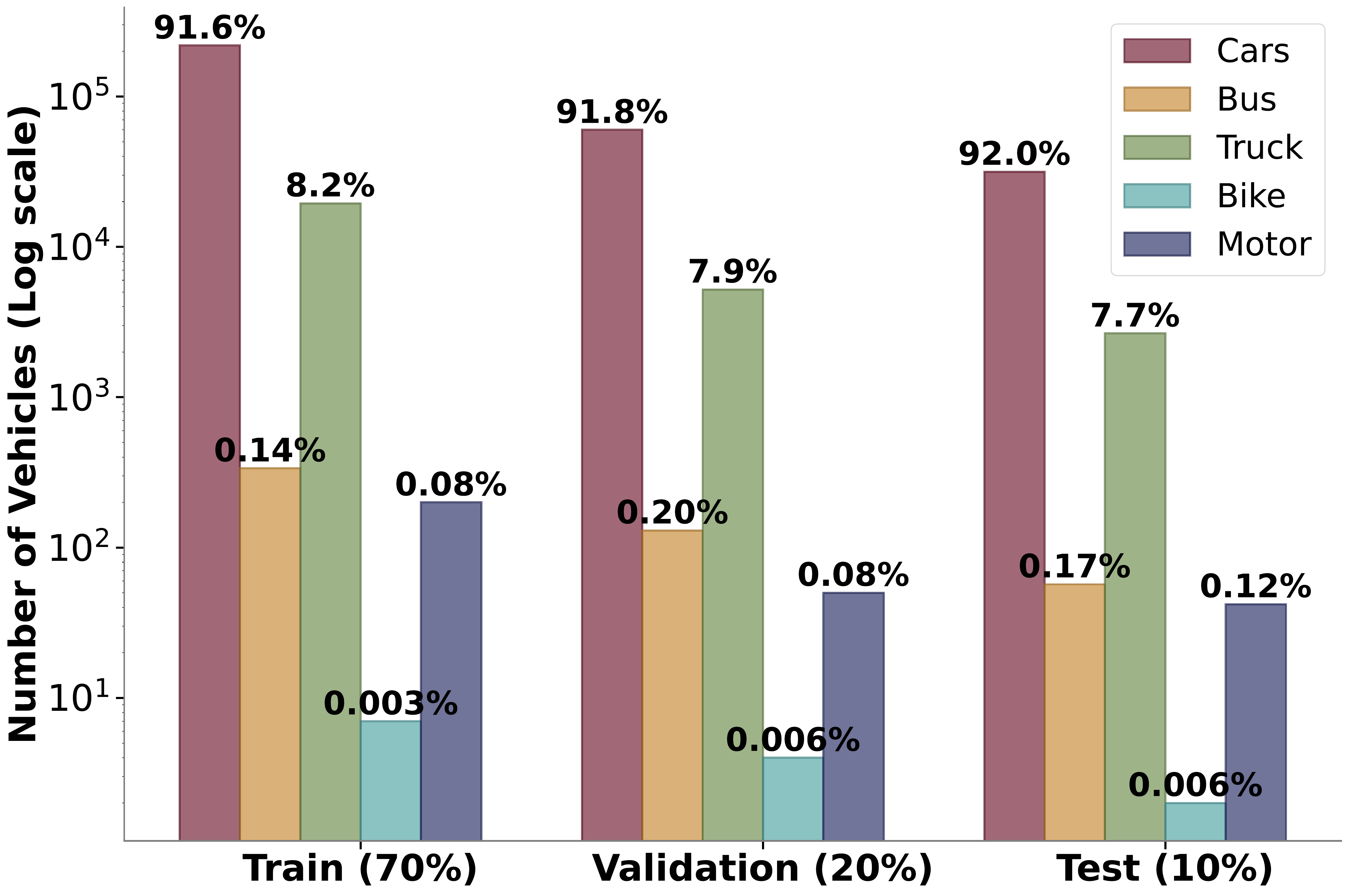}
\caption{Vehicle distribution in the CHD trajectory data across test, train, and validation sets.}
\vspace{-15pt}
\label{fig:graph_class}
\end{figure}

Ensuring consistency with commonly utilized trajectory prediction models \cite{katariya2022deeptrack1, noghre2022pishgu, li2019grip}, the trajectory data in CHD is extracted at a frame rate of 5 fps that was uniformly distributed with 70\% assigned to the training set, 20\% to the validation set, and 10\% to the test set. The trajectory data are also extracted at 60 fps to facilitate research at higher frame rates. This dataset consists of five different classes of vehicles, and the distribution of different classes across different sets is presented in Fig. \ref{fig:graph_class}. It can be seen that CHD has around 90\% cars, and the rest 9\% is divided between bus, truck, bike, and motor category.UA-DETRAC\cite{Wen2020UADETRAC} high-angle dataset from Tabel \ref{tab:cam_view}, exhibits similar distribution with around 87\% of cars and remaining 13\% distribution of other vehicles. Whereas, HighD\cite{Krajewski2018highD} and NGSIM\cite{NGSIM_i80, NGSIM_US101} trajectory datasets from Tabel \ref{tab:traj_summ}, has about 70\% and 96\% of cars.

CHD comprises several components for each recording, including raw videos, annotation data, and extracted trajectories. The annotation data file includes details such as the frame number, vehicle identification number, vehicle type, and bounding box coordinates. The vehicle trajectory data files contain frame and ID information and X and Y coordinates. Overall, the high frame rate, a large number of high-quality images and multiple POVs, and a comprehensive collection of real-world data make CHD a suitable benchmark for various ITS applications.


\section{PishguVe}
\label{sec:pishguVe}

Vehicle trajectory prediction aims to forecast where an individual vehicle will be in the future, T time steps ahead, based on their past positions. The proposed model considers intrinsic and extrinsic factors influencing a vehicle's trajectory to achieve high accuracy while being efficient in real time. Using observed trajectories of all vehicles in a given scene, the model can accurately predict future positions by incorporating temporal and spatial dependencies between the vehicles.

In this study, we distinguish various elements of vehicle trajectory prediction as follows: The past trajectories of vehicles are represented by a set of absolute coordinates, denoted as $V_{i}$, and a set of relative coordinates, denoted as $\Delta V_i$. The absolute coordinates are defined as $V_i = {(x_{i}^{t}, y_{i}^{t})}$, where ${t = 1, ..., T_{in}}$, ${i \in {1, 2, ..., n}}$ representing the index of the vehicle and ${x_{i}^{t}}$ and ${y_{i}^{t}}$ are x and y coordinates of the center of bounding box of vehicle ${i}$ at time ${t}$. The relative coordinates are defined as $\Delta V_i = {(x_{i}^{t} - x_{i}^{1}, y_{i}^{t} -y_{i}^{1})}$.

The study's objective is to predict the future trajectories of each vehicle, $\hat{Y}_i$, using its past trajectories as input. The predicted future trajectories are represented as $\hat{Y}_i = {(x_{i}^{t}, y_{i}^{t})}$, where ${t =  (T_{in}+1), ..., T_{out}}$ and ${i \in {1, 2, ..., n}}$, are generated as a set of coordinates for each vehicle, indicating their positions for future time steps. These predictions are evaluated by comparing them with the ground truth future trajectories, denoted as $Y$.

\begin{figure*}[]
\centering
\includegraphics[clip,trim={21 18 18 21},width=1\textwidth]{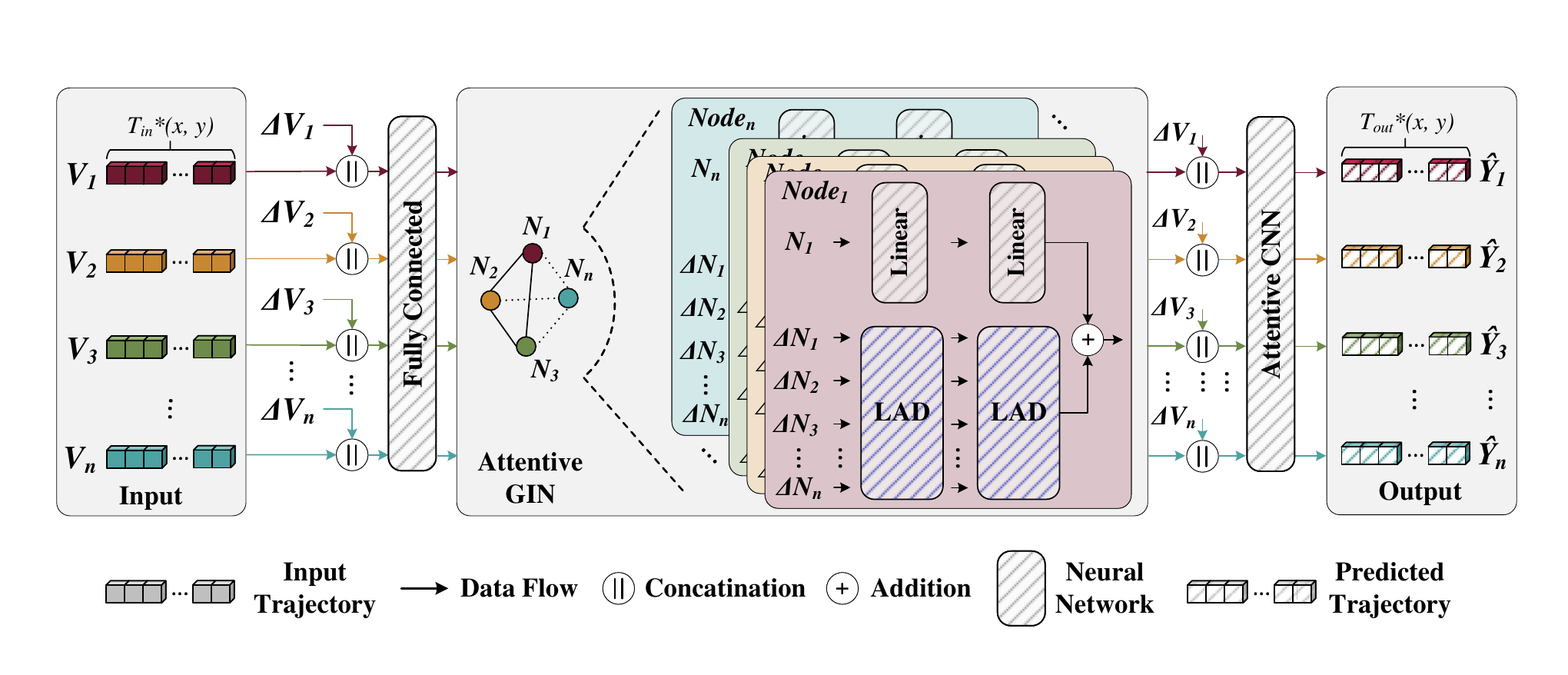}
\caption{ PishguVe architecture overview. Input trajectory vector, $V_{i}$ (where i $\in$ (1 to n)) and relative trajectory vector ,$\Delta V_{i}$ (where i $\in$ (1 to n)) with x and y coordinates and input time window of $T_{in}$ are concatenated and inferenced through the fully connected layer. The output of the fully connected layer, $N_{i}$, and neighbor aggregation vector $\Delta N_{i}$ is utilized by Attentive GIN in two separate branches for aggregating the node and neighbor features for detailed representation. The spatiotemporal attentive CNN is then used to predict the future trajectories of all vehicles up to $T_{out}$ time steps ahead.}
\vspace{-10pt}
\label{fig:system}
\end{figure*}

\subsection{Network Architecture}
\label{sec:architecture}
The overall architecture of Pishgu-Ve can be seen in Fig. \ref{fig:system}. In the first step, the absolute and the relative coordinates are concatenated together and passed through a fully connected layer to transform into the latent space:
\vspace{-10pt}

\begin{equation}
N_i = f\left(W_1.(V_i\mathbin\Vert \Delta V_i)+B_1\right)
\end{equation}
Where $N_i$, $f$, $W_1$, and $B_1$ represent the embedded representation of the $i^{th}$ vehicle in the scene, the Leaky ReLU activation function, and the corresponding weights and the biases of the fully connected layer accordingly. The absolute coordinates show the global position of each vehicle. In contrast, the relative coordinates provide information about how each vehicle moves with respect to its past trajectory, and the single fully connected layer integrates both into a latent representation. 

$N_i$ is then fed to the Graph Isomorphism Network (GIN) for capturing the interdependencies between available vehicles in the scene. Each vehicle is represented as a node in a fully connected graph. The graph is chosen to be fully connected to remove predefined biases and allow the network to decide how much data should be incorporated into the output in the message-passing process. On the other hand, having a fully connected graph enables PishguVe to combine features across all nodes in a single graph operation which helps with the efficiency of the architecture. Similar to \cite{noghre2022pishgu}, we adopt two separate networks for aggregating the node and neighbor features to create a richer presentation. For the node features, we utilize a fully connected network with one hidden layer: 
\vspace{-5pt}
\begin{equation}
G'_i = W_3.(W_2. (1+\theta) N_i+B_2) + B_3
\end{equation}

\begin{figure}[th]
\centering
\includegraphics[clip,trim={18 17 18 18}, width=1\linewidth]{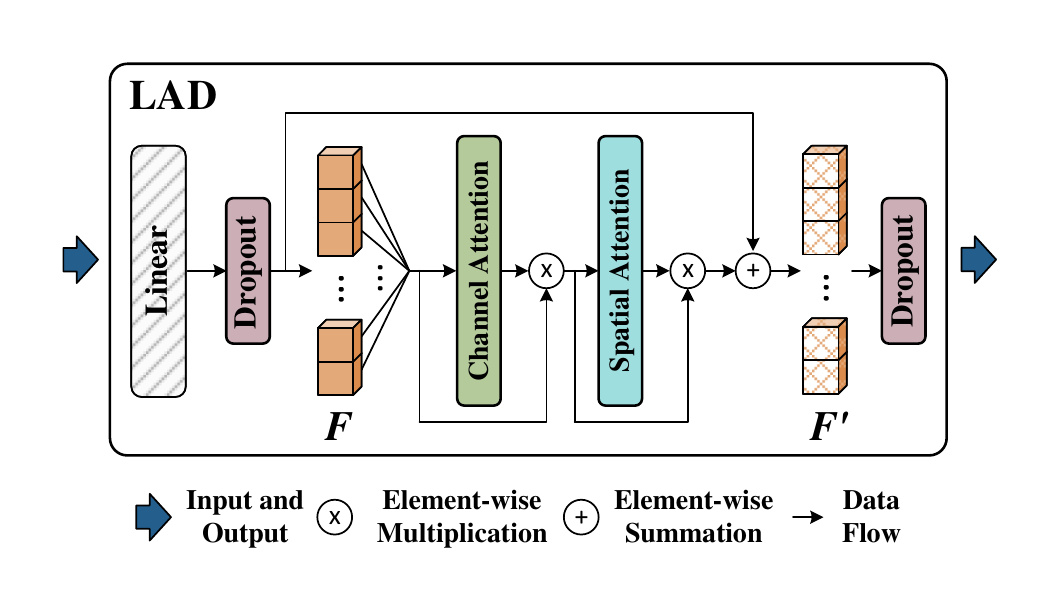}
\caption{The Linear Attention Dropout (LAD) Block. LAD is integrated into GIN for feature aggregation using channel and spatial attention. Dropout layers are added after linear and attention layers to minimize overfitting.}
\label{fig:lad}
\vspace{-15pt}
\end{figure}

Where $G'_i$ is the $i^{th}$ node aggregated features, $W_2$, $B_2$, $W_3$, and $B_3$ are the parameters of the two consecutive fully connected layers and $\theta$ is a learnable parameter. In contrast to \cite{noghre2022pishgu}, for neighbor aggregation, we define a new block called Linear Attention Dropout.

\textbf{Linear Attention Dropout} or LAD block is designed explicitly for neighbor feature aggregation. Fig. \ref{fig:lad} shows the details of the LAD block. LAD consists of a linear layer followed by attention layers for highlighting the most informative neighbor features to construct a richer representation. We adopt channel, and spatial attention from \cite{woo2018cbam} to integrate attention efficiently into the GIN. Also, to avoid overfitting, we leverage two dropout layers. Two back-to-back blocks of LAD are used for neighbor aggregation for extracting higher-level features. The LAD operation can be summarized as follows:
\begin{equation}
\begin{aligned}
F'_j = LAD(N_j) = & MLP_1(Pool_{avg}(W_4.\Delta N_j+B_4) +\\
& MLP_1(Pool_{max}(W_4.\Delta N_j+B_4))
\end{aligned}
\end{equation}

Where $W_4$ and $B_4$ are the parameters of the linear layer, $MLP_1$ is a shared Multi-layer Perceptron with one hidden layer, and $F'_j$ is the output of the LAD block. The final output of GIN will be constructed by the sum of outputs of the node aggregation and neighbor aggregation networks, which can be described as:
\begin{equation}
GIN(N_i) =G'_i + \sum_{j \in \mathcal{V}(i)} LAD(F'_j)
\end{equation}

Where $\mathcal{V}(i)$ represents the set of neighbors for the $i^{th}$ node.

In the next step, the enriched feature maps are concatenated with relative coordinates again to emphasize the relative movement compared to previous time steps and fed to an attentive CNN for the final trajectory prediction. CNNs have shown great capacity for extracting powerful feature maps. On top of that, we again use an efficient channel and spatial attention mechanism to improve the capability of finding more influential features. Keeping efficiency in mind, PishguVe has three convolutional layers followed by attention blocks and a last $1\times1$ convolutional layer for forming the final output. The first convolutional layer has a kernel size of $2\times2$ for capturing the low-frequency patterns, and the following two convolutional layers have a kernel size of $2\times1$.

\section{Evaluation and Experiments}
PishguVe is evaluated on the widely used NGSIM dataset \cite{NGSIM_US101,NGSIM_i80} and CHD dataset proposed in this manuscript. Similar to previous works\cite{deo2018convolutional,li2019grip, katariya2022deeptrack1,deo2018convolutional}, the 8 million data entries of the NGSIM were distributed 70\% training, 10\% validation, and 20\% testing sets. All models evaluated on CHD use a the data split discussed in Section \ref{subsec:stats}. All experiments were carried out on a Workstation equipped with a Threadripper Pro 3975WX processor with 32 cores clocking at 3.50 GHz and three A6000 GPUs.



\subsection{Evaluation Metrics}

\begin{figure}[h]
\centering
\includegraphics[width=0.5\textwidth]{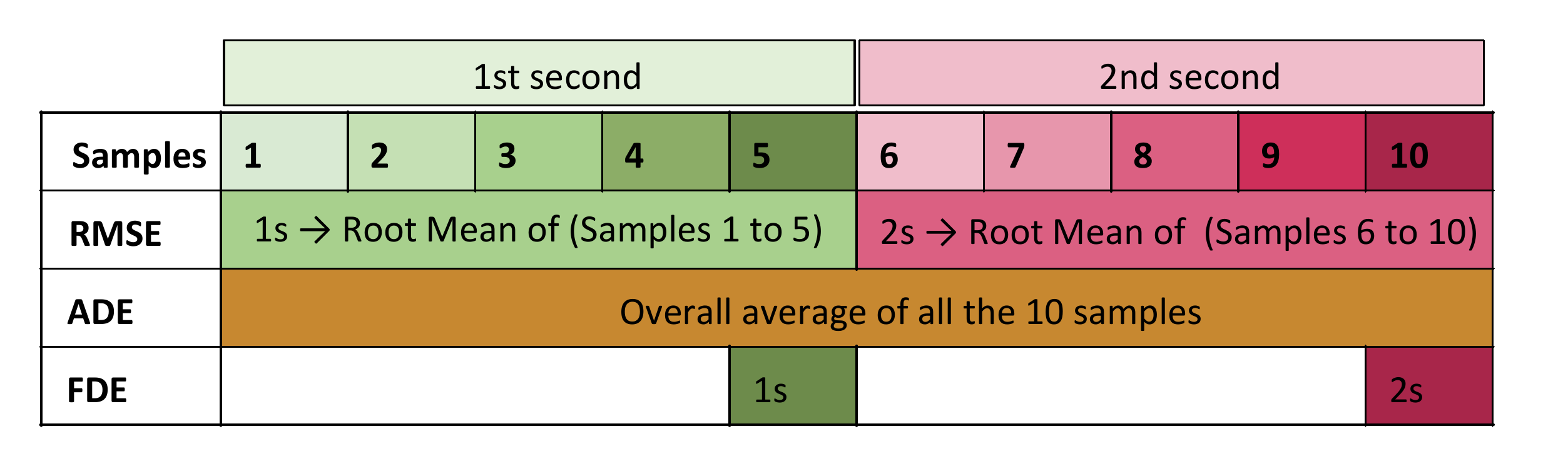}
\caption{Error matrices visualization for two seconds time window and a data rate of five samples/second.}
\label{fig:rmse}
\end{figure}

Root Mean Square Error (RMSE), Average Displacement Error (ADE), Final Displacement Error (FDE), and the number of model parameters are used as a measure of prediction accuracy and performance of the models. The error definitions are illustrated in Fig. \ref{fig:rmse}, visually representing the concepts under consideration. The figure assists in understanding each definition's nuances and highlighting their similarities and differences. 

\textbf{Root Mean Square Error (RMSE) }is commonly used to evaluate the accuracy of a predictive model that estimates the trajectories of vehicles in a scene. At a given time point, t, the RMSE is calculated as the square root of the mean square error between the predicted path ($\hat{Y}$) and the ground truth path ($Y$) of the ${n}$ subjects of interest in the scene:

\begin{equation}
\mathrm{RMSE}^{t} = \sqrt{\frac{1}{n}\sum_{i=1}^{n} (Y_{i}^{t} - \hat{Y}_{i}^{t})^{2}}
\end{equation}
\vspace{-15pt}

\begin{table*}[t]
\centering
\caption{Performance comparison of vehicle trajectory prediction approaches on NGSIM datasets\cite{NGSIM_i80, NGSIM_US101}. The best results are in bold letters, and the second to best are underlined.}
\begin{tabular}{c||c|c|c|c|c|c|c|c}
\rowcolor{DarkGray} {} & & & \multicolumn{5}{c|}{RMSE (m)} & \\
\rowcolor{Gray}
\textbf{Model} & \textbf{ADE   (m)} & \textbf{FDE   (m)} & \textbf{1s}   & \textbf{2s}   & \textbf{3s}   & \textbf{4s}   & \textbf{5s}  & \textbf{Params (K)}  \\ \hline \hline
CS-LSTM\cite{deo2018convolutional}       & 2.29                    & 3.34                    & 0.61          & 1.27          & 2.09          & 3.1           & 4.37       & 191   \\
GRIP++\cite{li2019grip}        & 2.01                    & 3.25                    & 0.38          & 0.89          & 1.45          & 2.14          & 2.94    & -      \\
STA-LSTM\cite{lin2021vehicle}        & 1.89                    & 3.16                    & 0.37          & 0.98          & 1.71          & 2.63          & 3.78   & \ul{124}      \\
DeepTrack\cite{katariya2022deeptrack1}    & 2.01                    & 3.21                    & 0.47          & 1.08          & 1.83          & 2.75          & 3.89  & \textbf{109}        \\
Pishgu\cite{noghre2022pishgu}         & \ul {0.88}              & \ul {1.96}              & \ul{0.15}    & \ul{0.46}    & \ul{0.82}    & \ul{1.25}          & \ul{1.74} & 132   \\ \\[-0.8em] \rowcolor{Gray} \hline
\textbf{PishguVe (Ours)} & \textbf{0.77}           & \textbf{1.76}           & \textbf{0.11} & \textbf{0.37} & \textbf{0.70} & \textbf{1.09} & \textbf{1.55} & 133.5 \\
\end{tabular}
\label{tab:ngsim_compare}
\end{table*}

\textbf{Average Displacement Error (ADE),} in the context of predicting the trajectories of vehicles in a scene, is an evaluation metric that measures the distance between the predicted ($\hat{Y}$) and the ground truth coordinates ($Y$) over all $T_{out}$ predicted time steps and all subjects of interest (${n}$) in the scene:

\begin{equation}
\mathrm{ADE} = \frac{1}{n * T_{out}} \sum_{i=1}^{n} \sum_{t=1}^{T_{out}} \left| Y_{i}^{t} - \hat{Y}_{i}^{t} \right|_{2}
\end{equation}

The $\left| \cdot \right|{2}$ notation denotes the L2 norm, which calculates the Euclidean distance between the predicted and actual coordinates. The ADE measures the average displacement error between the predicted and actual coordinates of the subjects of interest in the scene. A lower ADE indicates better accuracy of the predictive model.

\textbf{Final Displacement Error (FDE)} is an evaluation metric that measures the L2 distance between the predicted coordinates ($\hat{Y}$) and the ground truth coordinates ($Y$) of the last predicted time step for all subjects of interest (N) in the scene:

\begin{equation}
\mathrm{FDE} = \frac{1}{n} \sum_{i=1}^{n} \left| Y_{i}^{T_{out}} - \hat{Y}{i}^{T_{out}} \right|_{2}
\end{equation}

\textbf{Parameters} refer to the weights and biases of a neural network learned during training. The number of parameters affects models' performance and generalization ability.

\vspace{-10pt}
\subsection{Hyperparameter Search}
In this work, we extend the Graph Isomorphism Network (GIN) by adding attention mechanisms to improve performance. To evaluate the effectiveness of our approach, we conducted ablation studies and experiments such as adding attention at the node level in the Attentive GIN block, adding the LAD block at the node level, the relative nodes level, and both levels together. As expected, the best performance improvement was achieved by adding attention to the relative node level, which was missing a learnable parameter. However, overfitting was observed when the model was trained over a few epochs, as testing errors (i.e., ADE and FDE) kept rising with decreasing training errors. To overcome the overfitting, dropout layers were applied, as shown in Fig. \ref{fig:lad}. Table \ref{tab:ablation} shows the training ADE and FDE for PishguVe with different dropout probabilities. The tests were performed using the CHD High-Angle data and then extended to other datasets. Table \ref{tab:ablation} only shows the tests performed on the CHD High-Angle dataset, as it best represents the behavior of varying dropout values on PishguVe.



\begin{table}[b]
    \centering
    \caption{Average and final displacement error-based performance of PishguVe with varying dropout probabilities.}
    \begin{tabular}{c|c|c|c}
    
        \rowcolor{DarkGray}
        \textbf{Attention Dropout} & \textbf{Linear Dropout} & \textbf{ADE} & \textbf{FDE} \\ \hline \hline
        0.03 & 0.020 & 17.84 & 63.14 \\
        0.10 & 0.020 & 17.53 & 62.14 \\
        0.20 & 0.020 & 17.74 & 63.01 \\
        \rowcolor{Gray}
        0.25 & 0.020 & \textbf{16.81} & \textbf{57.71} \\
        0.40 & 0.020 & \ul{17.36} & 61.90 \\
        0.25 & 0.025 & 17.44 & \ul{61.46} \\
        0.25 & 0.15 & 17.86 & 62.22 \\
        0.40 & 0.15 & 17.82 & 62.59 \\
        0.40 & 0.30 & 18.78 & 68.51 \\ 
\end{tabular}
\label{tab:ablation}
\end{table}

The optimal dropout probabilities of the attention and linear layers were determined through experimentation, with the best ADE and FDE performance for probabilities of 0.25 and 0.02, respectively. The placement of dropout layers is shown in Fig. \ref{fig:lad}, and the performance of PishguVe with varying dropout probabilities is outlined in Table \ref{tab:ablation}.

All subsequent evaluations and comparisons of PishguVe presented in this manuscript were performed using the earlier probabilities for attention and linear dropout. These probabilities were selected to strike a balance between reducing overfitting and preserving the accuracy of the proposed model.




\subsection{Results}
This section evaluates the proposed PishguVe model against existing approaches on multiple datasets. Firstly, PishguVe is evaluated on the NGSIM dataset, including analysis and comparisons with existing approaches, as shown in Table \ref{tab:ngsim_compare}.

Next, performance PishguVe is also accessed on the proposed Carolinas Highway Dataset, with eye-level and high-angle POV data. This assessment excludes explicit models using lane number as one of the features as it is representative of real-world applications while still achieving best-in-class results.

\begin{table*}[t]
\centering
\caption{Performance comparison of vehicle trajectory prediction approaches on CHD eye-level POV data. The best results are in bold letters, and the second to best are underlined.}
\begin{tabular}{c||c|c|c|c|c|c|c|c}
\rowcolor{DarkGray} {} & & & \multicolumn{5}{c|}{RMSE (m)} & \\
\rowcolor{Gray}
\textbf{Model} & \textbf{ADE (pixels)} & \textbf{FDE (pixels)} & \textbf{1s}   & \textbf{2s}   & \textbf{3s}    & \textbf{4s}    & \textbf{5s} &\textbf{Params (K)}   \\ \hline \hline
Social-STGCNN\cite{mohamed2020social}  &  \ul{24.33}                   &     \ul{95.22}           &  \ul{4.32}             &   \ul{9.15}            &  \ul{15.93}              &   \ul{29.05}             & \ul{68.32}    & \textbf{7.4}     \\
GRIP++ \cite{li2019grip}        & 44.27                   & 129.58                   & 4.42        & 12.86         & 24.31         & 35.04 & 145.17    & -     \\
Pishgu   \cite{noghre2022pishgu}      & 37.99             & 123.69            & 4.98     & 13.58   & 26.61    & 50.31          & 106.45    & \ul{132}     \\ \\[-0.8em] \rowcolor{Gray} \hline
\textbf{PishguVe (Ours)}       & \textbf{20.75}           & \textbf{69.33}          & \textbf{3.21} & \textbf{8.24} & \textbf{15.55} & \textbf{28.46}    & \textbf{57.88} & 133.5
\end{tabular}
\label{tab:eye_level_comparision}
\end{table*}

\subsubsection{NGSIM-based trajectory prediction}
This section assesses the performance of PishguVe, by comparing its accuracy with several current vehicle trajectory prediction models on the NGSIM datasets \cite{NGSIM_i80, NGSIM_US101}. The first model we compare with is the Convolutional-Social LSTM (CS-LSTM)\cite{deo2018convolutional}, an encoder-decoder model that captures vehicle interactions to predict multiple trajectories for the ego vehicle. GRIP++\cite{li2019grip} is a graph-based approach that leverages vehicle-environment interactions to generate more accurate trajectory predictions. Another model we consider is the Spatiotemporal Attention-LSTM (STA-LSTM)\cite{lin2021vehicle}, which incorporates both spatial and temporal information and uses an attention mechanism to weigh the influence of historical trajectories and neighboring vehicles on the ego vehicle. The DeepTrack\cite{katariya2022deeptrack1} model is a temporal convolution networks-based encoder-decoder architecture that uses attention to predict a single trajectory for the ego vehicle. Finally, Pishgu\cite{noghre2022pishgu} is a GIN-based vehicle trajectory prediction model that employs an attentive CNN for prediction. The following sections present a thorough comparative analysis of PishguVe's performance against these models.

Table \ref{tab:ngsim_compare} shows that PishguVe performs better than all the selected models and achieves the lowest RMSE (at 1,2,3,4 and 5th seconds), ADE, and FDE values. However, DeepTrack has the least parameters, with STA-LSTM being the close second. PishguVe performs 12.50\% and 10.20\% better in ADE and FDE than the current SotA, Pishgu. It also performs better in terms of RMSE at each time step. Finally, we note that PishguVe has slightly more parameters (133.5K) than Pishgu (132K) but fewer parameters than CS-LSTM and GRIP++. This indicates that PishguVe achieves superior performance with a reasonable number of model parameters, making it an attractive choice for practical applications. PishguVe also achieves a minor but significant improvement over the Pishgu model. These results highlight the effectiveness of the proposed PishguVe architecture for vehicle trajectory prediction on the NGSIM dataset.

\begin{table*}[b]

\centering
\caption{Performance comparison of vehicle trajectory prediction approaches on CHD high-angle POV data. The best results are in bold letters, and the second to best are underlined.}
\begin{tabular}{c||c|c|c|c|c|c|c|c}
 \rowcolor{DarkGray} {} & & & \multicolumn{5}{c|}{RMSE (m)}  &        \\ 
\rowcolor{Gray}
\textbf{Model} & \textbf{ADE   (pixels)} & \textbf{FDE   (pixels)} & \textbf{1s}   & \textbf{2s}   & \textbf{3s}    & \textbf{4s}    & \textbf{5s}   &\textbf{Params (K)}  \\ \hline \hline
Social-STGCNN \cite{mohamed2020social} & 31.87                   & 98.46                   & 9.74         & 21.83         & 29.01         & 42.34 & 82.14      & 7.4    \\
GRIP++  \cite{li2019grip}       & 36.32                   & 100.89                   & \textbf{3.40}         & \textbf{6.67}        & 14.32        & 28.02 & 123.04   & -       \\
Pishgu  \cite{noghre2022pishgu}       & \ul {18.33}             & \ul{61.92}             & 4.04    & 7.48    & \ul{13.99}    & \ul{24.30}          & \ul{51.51}  &132       \\ \\[-0.8em] \rowcolor{Gray} \hline
\textbf{PishguVe (Ours)}      & \textbf{16.81}           & \textbf{57.71}          & \ul{3.52} & \ul{7.12} & \textbf{12.64} & \textbf{22.93}    & \textbf{48.60} &133.5
\end{tabular}
\label{tab:high_angle_comparision}
\end{table*}

\subsubsection{CHD for trajectory prediction pov}    
For trajectory prediction on the CHD datasets, we evaluated the performance of PishguVe against three contemporary models that do not utilize lane number as an input feature, GRIP++ \cite{li2019grip}, Pishgu \cite{noghre2022pishgu}, and Social-STGCNN \cite{mohamed2020social}. The exclusion of lane number as a feature is motivated by challenges faced by real-world systems, such as inaccuracies and errors in lane detection propagating to system output and sensitivity of lane detection approaches to external factors such as weather.

GRIP++ and Pishgu were also used for comparisons on the NGSIM dataset. The Social-STGCNN is included here due to its extremely low complexity and best-in-class performance in pedestrian trajectory prediction. It is used for modeling the spatiotemporal patterns of human movements in social groups. Each model was trained multiple times for several epochs on eye-level and high-angle datasets to obtain the best results. By comparing PishguVe's performance against these models, we aim to demonstrate the effectiveness of the proposed model in improving trajectory prediction accuracy while reducing the reliance on lane identification.

\textbf{CHD Eye-Level POV.} Table \ref{tab:eye_level_comparision} outlines the performance of the aforementioned models on the CHD Eye-Level dataset. PishguVe outperforms all other models with an ADE of 20.75 pixels and an FDE of 69.33 pixels, respectively. Specifically, compared to the second-best model, Social-STGCNN, PishguVe achieves a 14.58\% lower ADE and a 27.38\% lower FDE. 

Regarding prediction accuracy at different time horizons, PishguVe outperforms all other models, achieving the lowest RMSE values. Compared to Social-STGCNN achieves 25.74\%, 10.09\%, 2.40\%, 18.48\%, and 15.85\% lower RMSE values at 1s, 2s, 3s, 4s, and 5s, respectively. These results suggest that PishguVe is more accurate and effective for predicting vehicle paths than other SotA models for eye-level and birds-eye-view datasets. However, PishguVe reports the highest number of parameters, 133.5K, marginally behind the second-best model Pishgu with 132K parameters.

\textbf{CHD High-Angle POV.} The proposed PishguVe model was also trained on the CHD High-Angle dataset, and the performance results presented in Table \ref{tab:high_angle_comparision} were obtained using this dataset. PishguVe achieves an ADE of 16.81 pixels and an FDE of 57.71 pixels, which are lower than the other models. The second best model for ADE and FDE is Pishgu, with an ADE of 18.33 pixels and an FDE of 61.92 pixels. PishguVe achieves an 8.3\% improvement in ADE and a 6.9\% improvement in FDE over the second-best model. 

PishguVe also performs well in terms of RMSE for certain time thresholds. It achieves the lowest error rates for 3s, 4s, and 5s and is slightly lower than GRIP++ for 1s and 2s. Compared to the second-best model, Pishgu, PishguVe achieves a 13.9\%, 5.1\%, 0.7\%, 9.6\%, and 5.8\% improvement for 1s, 2s, 3s, 4s, and 5s, respectively.

Furthermore, PishguVe outperforms the Social-STGCNN model, which has relatively fewer parameters but struggles in generalizing results. The ADE and FDE performance of PishguVe in the High-Angle POV dataset is 89.36\% and 70.94\%, respectively. Moreover, for root mean square error (RMSE) comparisons to Social-STGCNN, PishguVe performs better by 178.29\%, 206.52\%, 129.10\%, 84.74\%, and 68.99\% for 1s, 2s, 3s, 4s, and 5s, respectively.

\section{Conclusion}


This paper presents the \textit{Carolinas Highway dataset (CHD)}, which consists of over 338,000 vehicle trajectories captured from 1.6M high-resolution images recorded at eight highway locations from two distinct POVs. The proposed dataset provides a unique benchmark for evaluating trajectory prediction and various highway-based applications in ITS. 

We also introduce \textit{PishguVe}, a SotA trajectory prediction architecture that utilizes graph-based attention and an attentive neural network to extract essential features and produce real-time results. It creates graphs of all vehicles in a scene to learn their dependency and behavior in different driving conditions. Our experiments on the NGSIM and CHD have demonstrated that PishguVe sets a new state-of-the-art for vehicle trajectory prediction in bird's-eye, high-angle, and eye-level POV, achieving better performance compared to existing approaches. The proposed architecture and dataset can facilitate the development of more advanced driver safety and assistance systems, intelligent transportation systems, and traffic analysis and surveillance tools, significantly improving road safety and efficiency.



\section*{Acknowledgements}
This work was supported by the National Science Foundation (NSF) under Award No. 1932524.


\bibliographystyle{IEEEtran}
\bibliography{Bibliography/main}

\section*{Biography}
\vspace{-20pt}
\begin{IEEEbiography}[{\includegraphics[width=1in,height=1.15in,clip,keepaspectratio]{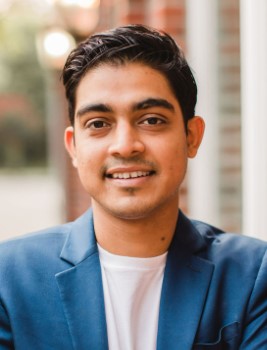}}]{Vinit Katariya} (S'14) is a PhD researcher working under the guidance of Dr. Hamed Tabkhi at Department of Electrical and Computer Engineering, University of North Carolina at Charlotte, USA. He received his Masters degree University of North Carolina at Charlotte in  2016. His current research interests include real-time Machine learning, Deep learning, Intelligent Transportation and Motion Forecasting. He is an active student member of IEEE.
\end{IEEEbiography}

\vspace{-40pt}

\begin{IEEEbiography}[{\includegraphics[width=1in,height=1.1in,keepaspectratio]{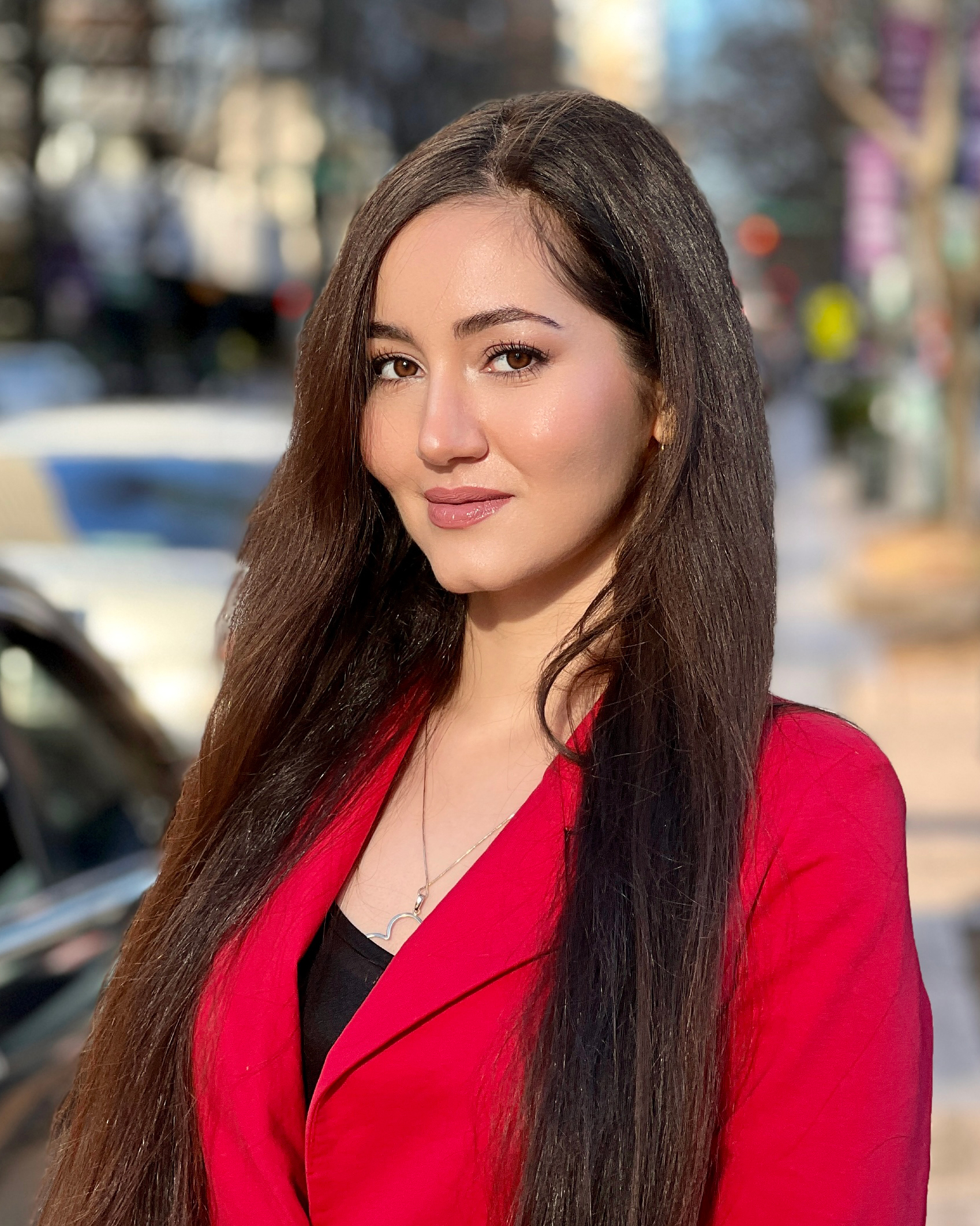}}]{Ghazal Alinezhad Noghre} (S’22) is currently pursuing her Ph.D. in Electrical and Computer Engineering at the University of North Carolina at Charlotte, NC, United States. Her research concentrates on Artificial Intelligence, Machine Learning, and Computer Vision. She is particularly interested in the applications of anomaly detection, action recognition, and path prediction in real-world environments, and the challenges associated with these fields.
\end{IEEEbiography}

\vspace{-40pt}

\begin{IEEEbiography}[{\includegraphics[width=1in,height=1.12in,keepaspectratio]{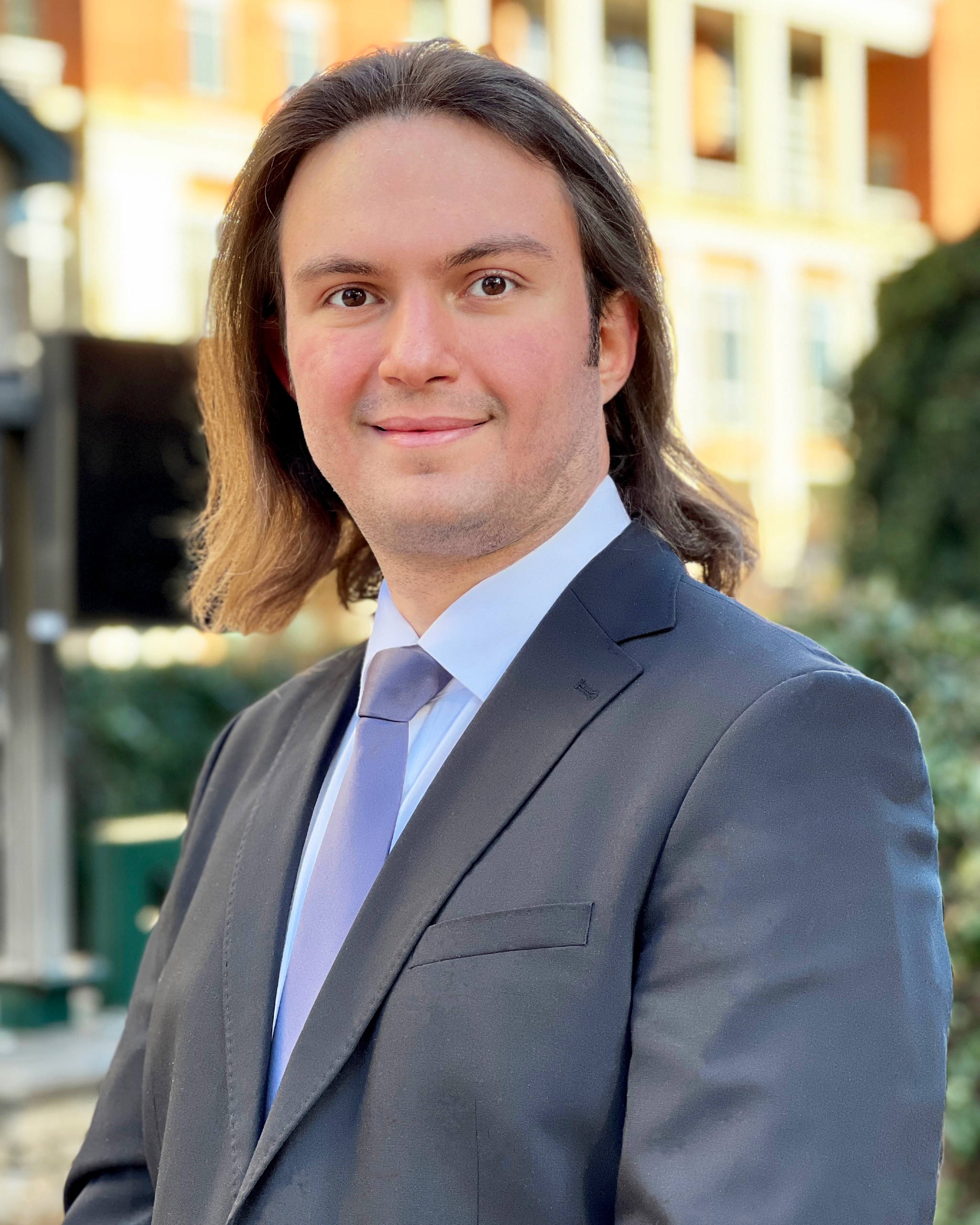}}]{Armin Danesh Pazho} (S’22) is currently a Ph.D. student at the University of North Carolina at Charlotte, NC, United States. With a focus on Artificial Intelligence, Computer Vision, and Deep Learning, his research delves into the realm of developing AI for practical, real-world applications and addressing the challenges and requirements inherent in these fields. Specifically, his research covers action recognition, anomaly detection, person re-identification, human pose estimation, and path prediction.
\end{IEEEbiography}

\vspace{-40pt}

\begin{IEEEbiography}[{\includegraphics[width=1in,height=1.25in,keepaspectratio]{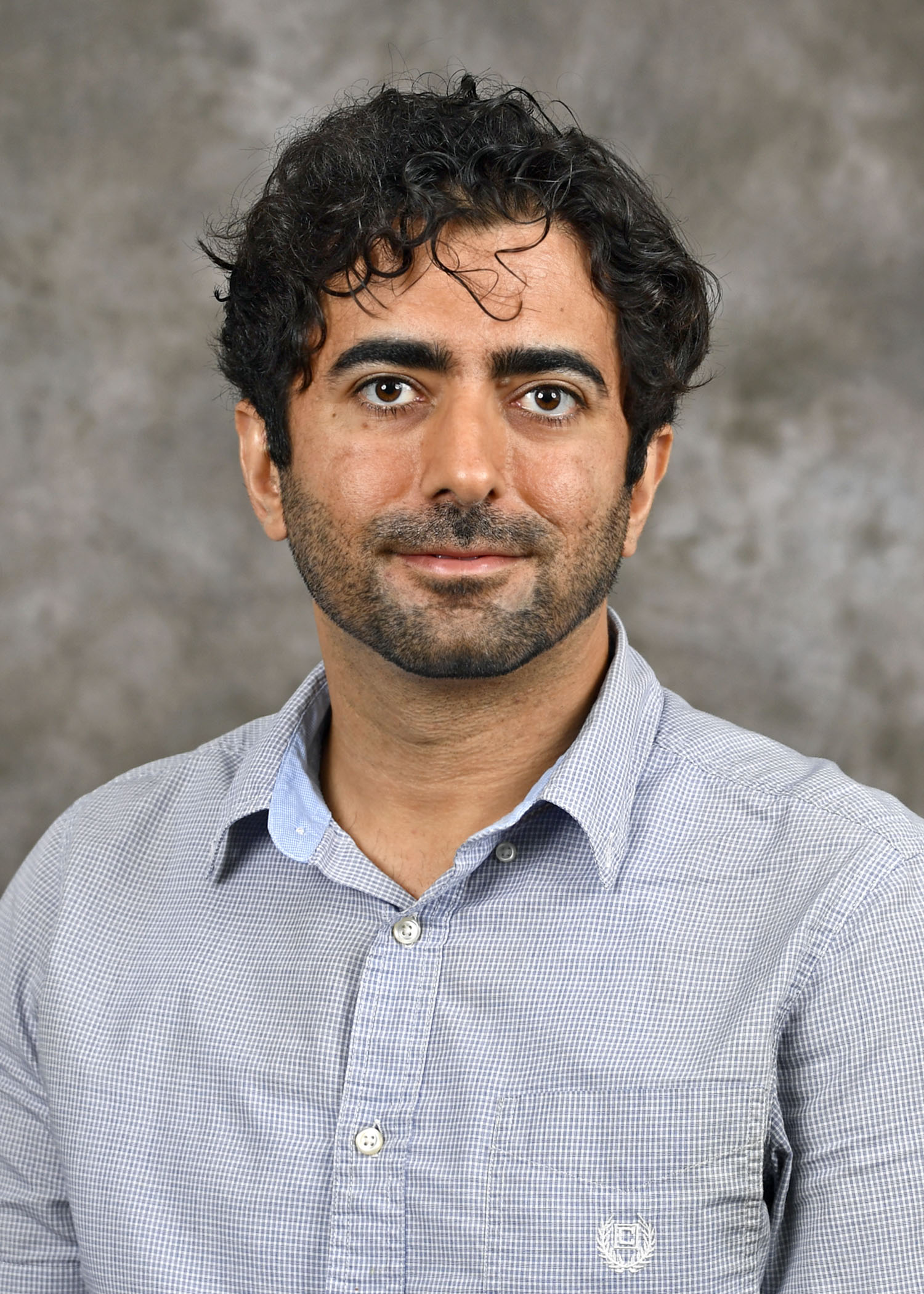}}]{Hamed Tabkhi}
(S’07–M’14)
is an Associate Professor in the Department of Electrical and Computer Engineering, University of North Carolina at Charlotte, USA.
He was a post-doctoral research associate at Northeastern University. Hamed Tabkhi received his Ph.D. degree in 2014 from Northeastern University under the direction of Prof. Gunar Schirner. Overall, his research focuses on transformative computer systems and architecture for cyber-physical, real-time streaming and emerging machine learning applications.
\end{IEEEbiography}

\vfill

\end{document}